\documentclass[sigplan,dvipsnames,10pt]{acmart}

\acmConference[PL'17]{ACM SIGPLAN Conference on Programming Languages}{January 01--03, 2017}{New York, NY, USA}
\acmYear{2017}
\acmISBN{} 
\acmDOI{} 
\startPage{1}

\setcopyright{none}

\usepackage[utf8]{inputenc}
\usepackage{natbib}
\usepackage{graphicx}
\usepackage{listings}
\usepackage{amsmath}
\usepackage{amssymb}
\usepackage{fullpage}
\usepackage{subcaption}
\usepackage{enumitem}
\usepackage{url}

\usepackage[nameinlink]{cleveref}
\creflabelformat{equation}{#2\textup{#1}#3}  

\newcommand{\m}[1]{\texttt{#1}}

\newcommand{\Exp}[1]{ {\operatorname{\sf E}\left[ #1 \right] } }

\definecolor{strings}{rgb}{.624,.251,.259}
\definecolor{keywords}{rgb}{.224,.451,.686}
\definecolor{comment}{rgb}{.322,.451,.322}

\lstdefinelanguage{python}{
  morekeywords={from, import, as, for, in, while, def, return, =, +, if, elif, else,
  -, /, *, lambda},
  keywords=[2]{1,2,3,4,5,6,7,8,9,0},
  keywords=[3]{list,len},
  morecomment=[l]{\#},
  morecomment=[s]{"""}{"""},
  morestring=[b]',
  morestring=[b]",
  alsoletter={<>=-+/*},
  sensitive=true
}

\newcommand{\specialcell}[2][t]{%
  \begin{tabular}[#1]{@{}l@{}}#2\end{tabular}}

\newcommand{\iid}{\mbox{$\;\stackrel{\tiny\rm iid}{\sim}\;$}}

\usepackage{ragged2e}

\begin{document}

\title{TensorFlow Distributions}

\author{Joshua V. Dillon$^*$, Ian Langmore$^*$, Dustin Tran$^{*\dagger}$, Eugene Brevdo$^*$, Srinivas Vasudevan$^*$, Dave Moore$^*$, Brian Patton$^*$, Alex Alemi$^*$, Matt Hoffman$^*$, Rif A. Saurous$^*$}
\affiliation{
  \institution{$^*$Google, $^\dagger$Columbia University}
}

\begin{abstract}
The TensorFlow Distributions library implements a vision of
probability theory adapted to the modern deep-learning paradigm of
end-to-end differentiable computation. Building on two basic
abstractions, it offers flexible building blocks for probabilistic
computation. \\ \m{Distribution}s provide fast, numerically stable
methods for generating samples and computing statistics, e.g., log
density.  \m{Bijector}s provide composable volume-\\tracking transformations with automatic caching. Together these enable modular construction of high dimensional distributions and transformations not possible with previous libraries (e.g., pixelCNNs, autoregressive flows, and reversible residual networks). They are the workhorse behind deep probabilistic programming systems like Edward and empower fast black-box inference in probabilistic models built on deep-network components. TensorFlow Distributions has proven an important part of the TensorFlow toolkit within Google and in the broader deep learning community.
\end{abstract}


\maketitle
\section{Introduction}
\label{sec:introduction}

The success of deep learning---and in particular, deep generative models---presents exciting opportunities for probabilistic programming. Innovations with deep probabilistic models and inference algorithms have enabled new successes in perceptual domains such as images \citep{goodfellow2014generative}, text \citep{anonymous2017generative}, and audio \citep{vandenoord2016wavenet}; and they have  advanced scientific applications such as understanding mouse behavior \citep{johnson2016composing}, learning causal factors in genomics \citep{tran2017implicit}, and synthesizing new drugs and materials \citep{gomez2016automatic}.

Reifying these applications in code falls naturally under the scope of probabilistic programming systems, systems which build and manipulate computable probability distributions. Within the past year, languages for deep probabilistic programming such as Edward \citep{tran2017deep} have expanded deep-learning research
by enabling new forms of experimentation, faster iteration cycles,
and improved reproducibility.

While there have been many developments in probabilistic programming languages, there has been limited progress in backend systems for probability distributions. This is despite their fundamental necessity for computing log-densities, sampling, and statistics, as well as for manipulations when composed as part of probabilistic programs. Existing distributions libraries lack modern tools necessary for deep probabilistic programming. Absent are:
batching, automatic differentiation,
GPU support, compiler optimization,
composability with numerical operations and higher-level modules such as neural network layers, and
efficient memory management.

\begin{figure}[tb!]
\begin{lstlisting}[basicstyle=\small\ttfamily]
e = make_encoder(x)
z = e.sample(n)
d = make_decoder(z)
r = make_prior()
avg_elbo_loss = tf.reduce_mean(
 e.log_prob(z) - d.log_prob(x) - r.log_prob(z))
train = tf.train.AdamOptimizer().minimize(
 avg_elbo_loss)
\end{lstlisting}
\vspace{-2ex}
\caption{General pattern for training a variational auto-encoder (VAE) \citep{kingma2014auto}.}
\label{fig:generic_vae}
\vspace{-3ex}
\end{figure}

\begin{figure}[tb!]
\begin{lstlisting}[basicstyle=\small\ttfamily]
def make_encoder(x, z_size=8):
  net = make_nn(x, z_size*2)
  return tfd.MultivariateNormalDiag(
    loc=net[..., :z_size],
    scale=tf.nn.softplus(net[..., z_size:])))

def make_decoder(z, x_shape=(28, 28, 1)):
  net = make_nn(z, tf.reduce_prod(x_shape))
  logits = tf.reshape(
    net, tf.concat([[-1], x_shape], axis=0))
  return tfd.Independent(tfd.Bernoulli(logits))

def make_prior(z_size=8, dtype=tf.float32):
  return tfd.MultivariateNormalDiag(
    loc=tf.zeros(z_size, dtype)))

def make_nn(x, out_size, hidden_size=(128,64)):
  net = tf.flatten(x)
  for h in hidden_size:
    net = tf.layers.dense(
      net, h, activation=tf.nn.relu)
  return tf.layers.dense(net, out_size)
\end{lstlisting}
\vspace{-2ex}
\caption{Standard VAE on MNIST with mean-field Gaussian encoder, Gaussian prior, Bernoulli decoder.}
\label{fig:simple_vae}
\vspace{-3.5ex}
\end{figure}

\begin{figure}[tb!]
\begin{lstlisting}[basicstyle=\small\ttfamily]
import convnet, pixelcnnpp

def make_encoder(x, z_size=8):
  net = convnet(x, z_size*2)
  return make_arflow(
    tfd.MultivariateNormalDiag(
      loc=net[..., :z_size],
      scale_diag=net[..., z_size:])),
    invert=True)

def make_decoder(z, x_shape=(28, 28, 1)):
  def _logit_func(features):
    # implement single autoregressive step,
    # combining observed features with
    # conditioning information in z.
    cond = tf.layers.dense(z,
      tf.reduce_prod(x_shape))
    cond = tf.reshape(cond, features.shape)
    logits = pixelcnnpp(
      tf.concat((features, cond), -1))
    return logits
  logit_template = tf.make_template(
    "pixelcnn++", _logit_func)
  make_dist = lambda x: tfd.Independent(
    tfd.Bernoulli(logit_template(x)))
  return tfd.Autoregressive(
    make_dist, tf.reduce_prod(x_shape))

def make_prior(z_size=8, dtype=tf.float32):
  return make_arflow(
    tfd.MultivariateNormalDiag(
      loc=tf.zeros([z_size], dtype)))

def make_arflow(z_dist, n_flows=4,
   hidden_size=(640,)*3, invert=False):
  maybe_invert = tfb.Invert if invert else tfb.Identity
  chain = list(itertools.chain.from_iterable([
    maybe_invert(tfb.MaskedAutoregressiveFlow(
      shift_and_log_scale_fn=tfb.\
      masked_autoregressive_default_template(
          hidden_size))),
    tfb.Permute(np.random.permutation(n_z)),
  ] for _ in range(n_flows)))
  return tfd.TransformedDistribution(
    distribution=z_dist,
    bijector=tfb.Chain(chain[:-1]))
\end{lstlisting}
\vspace{-2ex}
\caption{State-of-the-art architecture. It uses a PixelCNN++ decoder \citep{salimans2017pixelcnn++} and autoregressive flows \citep{kingma2016improved,papamakarios2017masked} for encoder and prior.
}
\label{fig:power_vae}
\vspace{-2ex}
\end{figure}

To this end, we describe TensorFlow Distributions (r1.4), a TensorFlow (TF) library offering efficient, composable manipulation of probability distributions.\footnote{%
Home: \href{https://www.tensorflow.org}{tensorflow.org};
Source: \href{https://github.com/tensorflow/tensorflow}{github.com/tensorflow/tensorflow}.
}

\textbf{Illustration.}
\Cref{fig:generic_vae} presents a template for a variational autoencoder under the TensorFlow Python API;%
\footnote{Namespaces: \m{tf=tensorflow}; \m{tfd=tf.contrib.distributions}; \m{tfb=tf.contrib.distributions.bijectors}.} this is a generative model of binarized MNIST digits trained using amortized variational inference \citep{kingma2014auto}.
\Cref{fig:simple_vae} implements a standard version with a Bernoulli decoder, fully factorized Gaussian encoder, and Gaussian prior. By changing a few lines, \Cref{fig:power_vae} implements a state-of-the-art architecture: a PixelCNN++ \citep{salimans2017pixelcnn++} decoder and a convolutional encoder and prior pushed through autoregressive flows \citep{kingma2016improved,papamakarios2017masked}. (\m{convnet,pixelcnnpp} are omitted for space.)

\Cref{fig:generic_vae,fig:simple_vae,fig:power_vae} demonstrate the power of TensorFlow Distributions: fast, idiomatic modules are composed to express rich, deep structure.
\Cref{sec:applications} demonstrates more applications: kernel density estimators, pixelCNN as a fully-visible likelihood, and how TensorFlow Distributions is used within higher-level abstractions (Edward and TF Estimator).

\textbf{Contributions.}
TensorFlow Distributions (r1.4) defines two abstractions: \m{Distribution}s and \m{Bijector}s. \m{Distribution}s provides a collection of
56
distributions with fast, numerically stable methods for sampling, computing log densities, and many statistics. \m{Bijector}s provides a collection of
22
composable transformations with efficient volume-tracking and caching.

TensorFlow Distributions is integrated with the TensorFlow ecosystem \citep{abadi2016tensorflow}:
for example,
it is compatible with
\texttt{tf.layers} for neural net architectures,
\texttt{tf.data} for data pipelines,
TF Serving for distributed computing, and TensorBoard for visualizations.
As part of the ecosystem, TensorFlow Distributions inherits and maintains integration with TensorFlow graph operations, automatic differentiation,
idiomatic batching and vectorization,
device-specific kernel optimizations and XLA, and accelerator support for CPUs, GPUs, and tensor processing units (TPUs) \citep{jouppi2017indatacenter}.

TensorFlow Distributions is widely used in diverse applications. It is used by production systems within Google and by Google Brain and DeepMind for research prototypes. It is the backend for Edward \citep{tran2016edward}.

\subsection{Goals}

TensorFlow Distributions is designed with three goals:

\textbf{Fast.}
TensorFlow Distributions is computationally and memory efficient.
For example, it strives to use only XLA-compatible ops (which enable compiler optimizations and portability to mobile devices), and whenever possible it uses differentiable ops (to enable end-to-end automatic differentiation).
Random number generators for sampling call device-specific kernels implemented in C++.
Functions with \m{Tensor} inputs also exploit vectorization through batches (\Cref{sub:shape}).
Multivariate distributions may be able to exploit additional vectorization structure.

\textbf{Numerically Stable.}
All operations in TensorFlow Distributions are numerically stable
across half, single, and double floating-point precisions (as
TensorFlow \m{dtype}s: \m{tf.bfloat16} (truncated floating point),
\m{tf.float16}, \\ \m{tf.float32}, \m{tf.float64}). Class constructors have a \m{validate\_args} flag for numerical asserts.

\textbf{Idiomatic.}
As part of the TensorFlow ecosystem, TensorFlow Distributions maintains idioms such as
inputs and outputs following a ``\m{Tensor}-in, \m{Tensor}-out'' pattern (though deviations exist; see \Cref{sub:higher}), outputs preserving the inputs' \m{dtype}, and preferring statically determined shapes.
Similarly, TensorFlow Distributions has no library dependencies besides NumPy \citep{walt2011numpy} and six \citep{petersonsix},
further manages \m{dtype}s, supports TF-style broadcasting, and simplifies shape manipulation.



\subsection{Non-Goals}

TensorFlow Distributions does not cover all use-cases. Here we highlight goals common to probabilistic programming languages which are specifically not goals of this library.\footnote{%
Users can subclass \m{Distribution} relaxing these properties.}

\textbf{Universality.}
In order to be fast, the \m{Distribution} abstraction makes an explicit restriction on the class of computable distributions. Namely, any \m{Distribution} should offer \m{sample} and \m{log\_prob} implementations that are computable in expected polynomial time.  For example, the Multinomial-LogisticNormal distribution \citep{blei2006correlated} fails to meet this contract.

This also precludes supporting a distributional calculus. For example, convolution is generally not analytic, so \m{Distribution}s do not support the \m{\_\_add\_\_} operator: if $X\sim f_X$, $Y\sim f_Y$, and share domain $D$, then $Z=X+Y$ implies $f_Z(z)=\int_D f_X(z-y)f_Y(y)dy=(f_X*f_Y)(z)$.\footnote{%
In future work, we may support this operation in cases when it satisfies our goals, e.g., for the analytic subset of stable distributions such as \m{Normal}, \m{Levy}.}

\textbf{Approximate Inference.}
\m{Distribution}s do not implement approximate inference or approximations of properties and statistics. For example, a Monte Carlo approximation of entropy is disallowed yet a function which computes an analytical, deterministic bound on entropy is allowed. Compound distributions with conjugate priors such as Dirichlet-Multinomial are allowed. The marginal distribution of a hidden Markov model is also allowed since hidden states can be efficiently collapsed with the forward-backward algorithm \citep{murphy2012machine}.

\section{Related Work}
\label{sec:related}

The R statistical computing language \citep{ihaka1996r}
provides a comprehensive collection of probability distributions. It inherits from the classic S language \citep{becker1984s} and has accumulated user contributions over decades.
We use R's collection as a goal for comprehensiveness and ease of user contribution.
TensorFlow Distributions differs in being object-oriented instead of functional, enabling manipulation of \m{Distribution} objects; operations are also designed to be fast and differentiable.
Most developers of the TensorFlow ecosystem are also Google-employed, meaning we benefit from more unification than R's ecosystem. For example, the popular \m{glmnet} and \m{lme4} R packages support only specific distributions for model-specific algorithms; all \m{Distribution}s support generic TensorFlow optimizers for training/testing.

The SciPy \texttt{stats} module in Python collects probability distributions and statistical functions \citep{jones2001scipy}.
TensorFlow's primary demographic is machine learning users and researchers; they typically use Python. Subsequently, we modelled our API after SciPy; this mimicks TensorFlow's API modelled after NumPy. Beyond API, the design details and implementations drastically differ. For example, TensorFlow Distributions
enables arbitrary tensor-dimensional vectorization,
builds operations in the TensorFlow computational graph,
supports automatic differentiation, and can run on accelerators.
The TensorFlow Distributions API also introduces innovations such as higher-order distributions (\Cref{sub:higher}), distribution functionals (\Cref{sub:functionals}), and \m{Bijector}s (\Cref{sec:bijector}).

Stan Math \citep{carpenter2015stan}
is a C++ templated library for numerical and statistical functions, and with automatic differentiation as the backend for the Stan probabilistic programming language \citep{carpenter2016stan}.
Different from Stan,
we focus on enabling deep probabilistic programming. This lead to new innovations with
bijectors,
shape semantics,
higher-order distributions,
and distribution functionals.
Computationally, TensorFlow Distributions also enables
a variety of non-CPU accelerators,
and compiler optimizations in static over dynamically executed graphs.

\section{\m{Distribution}s}

TensorFlow Distributions provides a collection of
approximately 60 distributions with fast, numerically stable methods for sampling, log density, and many statistics. We describe key properties and innovations below.

\vspace{-0.5ex}
\subsection{Constructor}
\label{sub:constructor}

TensorFlow \m{Distribution}s are object-oriented. A distribution implementation subclasses the \m{Distribution} base class. The base class
follows a ``public calls private'' design pattern where, e.g., the public \m{sample} method calls a private \m{\_sample} implemented by each subclass. This handles basic argument validation (e.g., type, shape) and simplifies sharing function documentation.

\m{Distribution}s take the following arguments:

\vspace{0.5ex}
\hspace{-1.5em}\begin{tabular}{ll}
\textbf{\m{parameters}}               & indexes to family \\
\textbf{\m{dtype}}                    & dtype of samples \\
\textbf{\m{reparameterization\_type}} & sampling (\Cref{sub:sampling}) \\
\textbf{\m{validate\_args}}           & \specialcell{bool permitting numerical \\ checks} \\
\textbf{\m{allow\_nan\_stats}}        & \specialcell{bool permitting NaN \\ outputs} \\
\textbf{\m{name}}                     & str prepended to TF ops
\end{tabular}
\vspace{0.5ex}

Parameter arguments support TF-style broadcasting. For example, \m{Normal(loc=1., scale=[0.5, 1., 1.5])} is effectively equivalent to \m{Normal(loc=[1., 1., 1.], scale=[0.5, 1., 1.5])}.
Distributions use self-\\documenting argument names from a concise lexicon. We never use Greek and prefer, for example, \m{loc}, \m{scale}, \m{rate}, \m{concentration}, rather than $\mu$, $\sigma$, $\lambda$, $\alpha$.

Alternative parameterizations can sometimes lead to an ``argument zoo.'' To migitate this, we distinguish between two cases. When numerical stability necessitates them, distributions permit mutually exclusive parameters (this produces only one extra argument). For example, \m{Bernoulli} accepts \m{logits} or \m{probs}, \m{Poisson} accepts \m{rate} or \m{log\_rate}; neither permits specifying both. When alternative parameterizations are structural, we specify different classes: \\ \m{MultivariateNormalTriL}, \m{MultivariateNormalDiag}, \\ \m{MultivariateNormalDiagPlusLowRank} implement multivariate normal distributions with different covariance structures.

The \m{dtype} defaults to floats or ints, depending on the distribution's support, and with precision given by its parameters'. Distributions over integer-valued support (e.g., \m{Poisson}) use \m{tf.int*}. Distributions over real-valued support (e.g., \m{Dirichlet}) use \m{tf.float*}. This distinction exists because of mathematical consistency; and in practice, integer-valued distributions are often used as indexes into \m{Tensor}s.\footnote{%
Currently, TensorFlow Distributions' \m{dtype} does not follow this standard. For backwards compatibility, we are in the progress of implementing it by adding a new \m{sample\_dtype} kwarg.
}

If \m{validate\_args=True}, argument validation happens during graph construction when possible; any validation at graph execution (runtime) is gated by a Boolean. Among other checks, \m{validate\_args=True} limits integer distributions' support to integers exactly representable by same-size IEEE754 floats, i.e., integers cannot exceed $2^\text{significand\_bits}$. If \m{allow\_nan\_stats=True}, operations can return NaNs; otherwise an error is raised.


\vspace{-0.5ex}
\subsection{Methods}

At minimum, supported \m{Distribution}s implement the following methods: \m{sample} to generate random outcome \m{Tensor}s, \m{log\_prob} to compute the natural logarithm of the probability density (or mass) function of random
outcome \m{Tensor}s, and \m{batch\_shape\_tensor}, \\ \m{event\_shape\_tensor} to describe the dimensions of random outcome \m{Tensor}s (\Cref{sub:shape}), returned itself as \m{Tensor}s.

Supported \m{Distribution}s implement many additional methods, including \m{cdf}, \m{survival\_function}, \m{quantile}, \m{mean}, \m{variance}, and \m{entropy}. The \m{Distribution} base class automates implementation of related functions such as \m{prob} given \m{log\_prob} and \m{log\_survival\_fn} given \m{log\_cdf} (unless a more efficient or numerically stable implementation is available).
Distribution-specific statistics are permitted; for example, \m{Wishart} implements the expected log determinant (\m{mean\_log\_det}) of matrix variates, which would not be meaningful for univariate distributions.

All methods of supported distributions satisfy the following contract:

\textbf{Efficiently Computable.}
All member functions have expected polynomial-time complexity. Further, they are vectorized (\Cref{sub:shape}) and have optimized sampling routines (\Cref{sub:sampling}).
TensorFlow Distributions also favors efficient parameterizations: for example, we favor \m{MultivariateNormalTriL}, whose covariance is parameterized by the outer product of a lower triangular matrix, over \m{MultivariateNormalFullCovariance} which requires a Cholesky factorization.

\textbf{Statistically Consistent.}
Under \m{sample}, the Monte Carlo approximation of any statistic converges to the statistic's implementation as the number of samples approaches $\infty$. Similarly, \m{pdf} is equal to the derivative of \m{cdf} with respect to its input; and \m{sample} is equal in distribution to uniform sampling followed by the inverse of \m{cdf}.

\textbf{Analytical.}
All member functions are analytical excluding \m{sample}, which is non-deterministic. For example, \m{Mixture} implements an analytic expression for an entropy lower bound method, \texttt{entropy\_lower\_bound}; its exact entropy is intractable. However, no method function's implementation uses a Monte Carlo estimate (even with a fixed seed, or low-discrepancy sequence \citep{niederreiter1978quasi})
which we qualify as non-analytical.


\textbf{Fixed Properties.}
In keeping with TensorFlow idioms,
\m{Distribution} instances have fixed shape semantics (\Cref{sub:shape}), \m{dtype}, class methods, and class properties throughout the instance's lifetime.
Member functions have no side effects other than to add ops to the TensorFlow graph.

Note this is unlike the statefulness of
exchangeable random primitives \citep{ackerman2016exchangeable}, where sampling can memoize over calls to lazily evaluate infinite-dimensional data structures. To handle such distributions, future work may involve a sampling method which returns another distribution storing those samples. This preserves immutability while enabling  marginal representations of completely random measures such as a Chinese restaurant process, which is the compound distribution of a Dirichlet process and multinomial distribution \citep{aldous1985exchangeability}. Namely, its \m{sample} computes a P\'{o}lya urn-like scheme caching the number of customers at each table.\footnote{%
While the Chinese restaurant process is admittable as a (sequence of) \m{Distribution}, the Dirichlet process is not: its probability mass function involves a countable summation.}

\subsection{Shape Semantics}
\label{sub:shape}

\begin{figure}
    \centering
    \footnotesize{
    \begin{equation*}\bigg[
    \underbrace{\parbox{2cm}{\raggedright $n$ Monte Carlo draws}}_{\parbox{2cm}{\raggedright\m{sample\_shape} (indep, identically distributed)}},
    \underbrace{\parbox{2cm}{\raggedright $b$ examples per batch}}_{\parbox{2cm}{\raggedright\m{batch\_shape} (indep, {\em not} identical)}},
    \underbrace{\parbox{2cm}{\raggedright $s$ latent dimensions}}_{\parbox{2cm}{\raggedright\m{event\_shape} (can be dependent)}} \bigg]
    \end{equation*}}
    \caption{Shape semantics. Refers to variational distribution in \Cref{fig:generic_vae}.}
    \label{fig:vae_shape}
\end{figure}

To make \m{Distribution}s fast, a key challenge is to enable arbitrary tensor-dimensional vectorization. This allows users to properly utilize multi-threaded computation as well as array data-paths in modern accelerators. However, probability distributions involve a number of notions for different dimensions; they are often conflated and thus difficult to vectorize.

To solve this, we (conceptually) partition a \m{Tensor}'s shape into three groups:
\begin{enumerate}
\item
{\em Sample shape} describes independent, identically distributed draws from the distribution.
\item
{\em Batch shape} describes independent, \emph{not} identically distributed draws. Namely, we may have a set of (different) parameterizations to the same distribution. This enables the common use case in machine learning of a ``batch'' of examples, each modelled by its own distribution.
\item
{\em Event shape} describes the shape of a single draw (event space) from the distribution; it may be dependent across dimensions.
\end{enumerate}
\Cref{fig:vae_shape} illustrates this partition for the latent code in a variational autoencoder (\Cref{fig:generic_vae}). Combining these three shapes in a single \m{Tensor} enables efficient, idiomatic vectorization and broadcasting.

Member functions all comply with the distribution's shape semantics and \texttt{dtype}. As another example, we initialize a batch of three multivariate normal distributions in $\mathbb{R}^2$. Each batch member has a different mean.
\begin{lstlisting}[basicstyle=\small\ttfamily]
# Initialize 3-batch of 2-variate
# MultivariateNormals each with different mean.
mvn = tfd.MultivariateNormalDiag(
  loc=[[1., 1.], [2., 2.], [3., 3.]]))
x = mvn.sample(10)
# ==> x.shape=[10, 3, 2]. 10 samples across
#     3 batch members. Each sample in R^2.
pdf = mvn.prob(x)
# ==> pdf.shape=[10, 3]. One pdf calculation
#     for 10 samples across 3 batch members.
\end{lstlisting}
Partitioning \m{Tensor} dimensions by ``sample'', ``batch'', and ``event'' dramatically simplifies user code while naturally exploiting vectorization. For example, we describe a Monte Carlo approximation of a Normal-Laplace compound distribution,
\begin{equation*}
\hspace{-0.5em}
p(x\mid\sigma,\mu_0,\sigma_0) = \int_\mathbb{R}
\operatorname{\sf Normal(x\mid\mu,\sigma)}
\operatorname{\sf Laplace}(\mu\mid\mu_0,\sigma_0)\,d\mu.
\end{equation*}
\begin{lstlisting}[basicstyle=\small\ttfamily]
# Draw n iid samples from a Laplace.
mu = tfd.Laplace(
  loc=mu0, scale=sigma0).sample(n)
# ==> sample_shape = [n]
#     batch_shape = []
#     event_shape = []
# Compute n different Normal pdfs at
# scalar x, one for each Laplace draw.
pr_x_given_mu = tfd.Normal(
  loc=mu, scale=sigma).prob(x)
# ==> sample_shape = []
#     batch_shape = [n]
#     event_shape = []
# Average across each Normal's pdf.
pr_x = tf.reduce_mean(pr_x_given_mu, axis=0)
# ==> pr_estimate.shape=x.shape=[]
\end{lstlisting}
This procedure is automatically vectorized because the internal calculations are over tensors, where each represents the differently parameterized \m{Normal} distributions. \m{sigma} and \m{x} are automatically broadcast; their value is applied pointwise thus eliding \m{n} copies.

To determine batch and event shapes (sample shape is determined from the \m{sample} method), we perform shape inference from parameters at construction time.
Parameter dimensions beyond that necessary for a single distribution instance always determine batch shape.
Inference of event shapes is typically not required as distributions often know it a priori; for example, \m{Normal} is univariate.
On the other hand, \m{Multinomial} infers its event shape from the rightmost dimension of its \m{logits} argument.
%
Dynamic sample and batch ranks are not allowed because they conflate the shape semantics (and thus efficient computation); dynamic event ranks are not allowed as a design choice for consistency.

Note that event shape (and shapes in general) reflects the numerical shape and not the mathematical definition of dimensionality. For example, \m{Categorical} has a scalar event shape over a finite set of integers; while a one-to-one mapping exists, \m{OneHotCategorical} has a vector event shape over one-hot vectors.
Other distributions with non-scalar event shape include Dirichlet (simplexes) and Wishart (positive semi-definite matrices).

\subsection{Sampling}
\label{sub:sampling}

Sampling is one of the most common applications of a \m{Distribution}. To optimize speed, we
implement sampling by registering device-specific kernels in C++ to TensorFlow operations. We also use well-established algorithms for random number generation.
For example, draws from \m{Normal} use the Box-Muller transform to return an independent pair of normal
samples from an independent pair of uniform samples \citep{boxmullertransform}; CPU, GPU, and TPU implementations exist.
Draws from \m{Gamma} use the rejection sampling algorithm of \citet{marsaglia2000simple}; currently, only a CPU implementation exists.

\textbf{Reparameterization.}
\m{Distribution}s employ a \\ \m{reparameterization\_type} property (\Cref{sub:constructor}) \\ which annotates the interaction between automatic differentiation and sampling.
Currently, there are two such annotations: ``fully reparameterized'' and ``not reparameterized''.
To illustrate ``fully reparameterized'', consider \m{dist = Normal(loc, scale)}. The sample \\\m{y = dist.sample()} is implemented internally via \m{x = tf.random\_normal([]); y = scale * x + loc}. The sample \m{y} is ``reparameterized'' because it is a smooth function of the parameters \m{loc}, \m{scale}, and a parameter-free sample \m{x}.
In contrast, the most common Gamma sampler is ``not reparameterized'': it uses an accept-reject scheme that makes the samples depend non-smoothly on the parameters \cite{naesseth-reparameterizable-rejection}.

When composed with other TensorFlow ops, a ``fully reparameterized'' \m{Distribution} enables end-to-end automatic differentiation on functions of its samples.  A common use case is a loss depending on expectations of the form $\Exp{\varphi(Y)}$ for some function $\varphi$. For example, variational inference algorithms minimize the KL divergence between $p_Y$ and another distribution $h$, \\ $\Exp{\log[p_Y(Y) / h(Y)]}$ using gradient-based optimization.
To do so, one can compute a Monte Carlo approximation
\begin{align}
    \label{align:finite-sample-estimate}
    S_N :&= \frac{1}{N} \sum_{n=1}^N \varphi(Y_n), \text{ where } Y_n\sim p_Y.
\end{align}
This lets us use $S_N$ not only as an estimate of our expected loss $\Exp{\varphi(Y)}$, but also use $\nabla_\lambda S_N$ as an estimate of the gradient $\nabla_\lambda \Exp{\varphi(Y)}$ with respect to parameters $\lambda$ of $p_Y$. If the samples $Y_n$ are reparameterized (in a smooth enough way), then both approximations are justified \citep{fu-gradient-estimation,schulman2015gradient,kingma2014auto}.


\subsection{Higher-Order Distributions}
\label{sub:higher}
Higher-order distributions are \m{Distribution}s which are functions of other \m{Distribution}s. This deviation from the \m{Tensor}-in, \m{Tensor}-out pattern enables modular, inherited construction of an enormous number of distributions. We outline three examples below and use a running illustration of composing distributions.

\textbf{\m{TransformedDistribution}} is a distribution $p(y)$ consisting of a base distribution $p(x)$ and invertible, differentiable transform $Y = g(X).$
The base distribution is an instance of the \m{Distribution} class and the transform is an instance of the \m{Bijector} class (\Cref{sec:bijector}).

For example, we can construct a (standard) Gumbel distribution from an Exponential distribution.
\begin{lstlisting}[basicstyle=\small\ttfamily, label=standard-gumbel]
standard_gumbel = tfd.TransformedDistribution(
  distribution=tfd.Exponential(rate=1.),
  bijector=tfb.Chain([
    tfb.Affine(
      scale_identity_multiplier=-1.,
      event_ndims=0),
    tfb.Invert(tfb.Exp()),
  ]))
standard_gumbel.batch_shape # ==> []
standard_gumbel.event_shape # ==> []
\end{lstlisting}
The \m{Invert(Exp)} transforms the Exponential distribution by the natural-log, and the \m{Affine} negates. In general, algebraic relationships of random variables are powerful, enabling distributions to inherit method implementations from parents (e.g., internally, we
implement multivariate normal distributions as \m{Affine} transforms of \m{Normal}).

Building on \m{standard\_gumbel}, we can also construct $2* 28* 28$ independent relaxations of the Bernoulli distribution, known as Gumbel-Softmax or Concrete \cite{maddison2017concrete,jang2017categorical}.
\begin{lstlisting}[basicstyle=\small\ttfamily]
alpha = tf.stack([
  tf.fill([28 * 28], 2.),
  tf.ones(28 * 28)])
concrete_pixel = tfd.TransformedDistribution(
  distribution=standard_gumbel,
  bijector=tfb.Chain([
    tfb.Sigmoid(),
    tfb.Affine(shift=tf.log(alpha)),
  ]),
  batch_shape=[2, 28 * 28])
concrete_pixel.batch_shape # ==> [2, 784]
concrete_pixel.event_shape # ==> []
\end{lstlisting}
The \m{Affine} shifts by \m{log(alpha)} for two batches. Applying \m{Sigmoid} renders a batch of $[2, 28* 28]$ univariate Concrete distributions.

\textbf{\m{Independent}}
enables idiomatic manipulations between batches and event dimensions.
Given a \m{Distribution} instance \m{dist} with batch dimensions, \m{Independent} builds a vector (or matrix, or tensor) valued distribution whose event components default to the rightmost batch dimension of \m{dist}.

Building on \m{concrete\_pixel}, we reinterpret the $784$ batches as jointly characterizing a distribution.
\begin{lstlisting}[basicstyle=\small\ttfamily]
image_dist = tfd.TransformedDistribution(
  distribution=tfd.Independent(concrete_pixel),
  bijector=tfb.Reshape(
    event_shape_out=[28, 28, 1],
    event_shape_in=[28 * 28]))
image_dist.batch_shape # ==> [2]
image_dist.event_shape # ==> [28, 28, 1]
\end{lstlisting}

The \m{image\_dist} distribution is over $28\times 28\times 1$-dim. events (e.g., MNIST-resolution pixel images).


\textbf{\m{Mixture}} defines a probability mass function $p(x) = \sum_{k=1}^K \pi_k p(x\,|\, k)$, where the mixing weights $\pi_k$ are provided by a \m{Categorical} distribution as input, and the components $p(x\,|\,k)$ are arbitrary \m{Distribution}s with same support. For components that are batches of the same family, \m{MixtureSameFamily} simplifies the construction where its components are from the rightmost batch dimension. Building on \m{image\_dist}:

\begin{lstlisting}[basicstyle=\small\ttfamily]
image_mixture = tfd.MixtureSameFamily(
  mixture_distribution=tfd.Categorical(
    probs=[0.2, 0.8]),
  components_distribution=image_dist)
image_mixture.batch_shape # ==> []
image_mixture.event_shape # ==> [28, 28, 1]
\end{lstlisting}
Here, \m{MixtureSameFamily} creates a mixture of two components with weights $[0.2, 0.8]$. The components are slices along the batch axis of \m{image\_dist}.



\subsection{Distribution Functionals}
\label{sub:functionals}

Functionals which take probability distribution(s) as input and return a scalar are ubiquitous.
They include information measures such as entropy, cross entropy, and mutual information \citep{cover1991elements}; divergence measures such as Kullback-Leibler, Csisz{\'a}r-Morimoto $f$-divergence \citep{csiszar1963informationstheoretische,morimoto1963markov}, and multi-distribution divergences \citep{garcia2012divergences}; and distance metrics such as integral probability metrics \citep{muller1997integral}.


We implement all (analytic) distribution functionals as methods in \m{Distribution}s.
For example, below we write functionals of Normal distributions:
\begin{lstlisting}
p = tfd.Normal(loc=0., scale=1.)
q = tfd.Normal(loc=-1., scale=2.)
xent = p.cross_entropy(q)
kl = p.kl_divergence(q)
# ==> xent - p.entropy()
\end{lstlisting}
\vspace{-1ex}


\section{\m{Bijector}s}
\label{sec:bijector}

We described \m{Distribution}s, sources of stochasticity which collect properties of probability distributions. \m{Bijector}s are deterministic transformations of random outcomes and of equal importance in the library. It consists of 22 composable transformations for manipulating \m{Distribution}s, with efficient volume-tracking and caching of pre-transformed samples. We describe key properties and innovations below.

\subsection{Motivation}
\label{sub:motivation}

The \m{Bijector} abstraction is motivated by two challenges for enabling efficient, composable manipulations of probability distributions:

\begin{enumerate}
\item
We seek a minimally invasive interface for manipulating distributions. Implementing every transformation of every distribution results in a combinatorial blowup and is not realistically maintainable. Such a policy is unlikely to keep pace with the pace of research. Lack of encapsulation exacerbates already complex ideas/code and discourages community contributions.
\item
In deep learning, rich high-dimensional densities typically use invertible volume-preserving mappings or mappings with fast volume adjustments (namely, the logarithm of the Jacobian's determinant has linear complexity with respect to dimensionality) \citep{papamakarios2017masked}. We'd like to efficiently and idiomatically support them.
\end{enumerate}

By isolating stochasticity from determinism, \m{Distribution}s are easy to design, implement, and validate.  As we illustrate with the flexibility of \m{TransformedDistribution} in \Cref{sub:higher}, the ability to simply swap out functions applied to the distribution is a surprisingly powerful asset.
Programmatically, the \m{Bijector} distinction enables encapsulation and modular distribution constructions with inherited, fast method implementations. Statistically, \m{Bijector}s enable exploiting algebraic relationships among random variables.

\subsection{Definition}
To address \Cref{sub:motivation},
the \m{Bijector} API provides an interface for transformations of distributions suitable for any differentiable and bijective map (\emph{diffeomorphism}) as well as certain non-injective maps (\Cref{subsection:smooth-covering}).

Formally, given a random variable $X$ and diffeomorphism $F$, we can define a new random variable $Y$ whose density can be written in terms of $X$'s,
\begin{align}
    \label{align:transform-pdf}
    p_Y(y) &= p_X(F^{-1}(y))~|DF^{-1}(y)|,
\end{align}
where $DF^{-1}$ is the inverse of the Jacobian of $F$.
Each \m{Bijector} subclass corresponds to such a function $F$, and \m{TransformedDistribution} uses the bijector to automate the details of the transformation $Y = F(X)$'s density (\Cref{align:transform-pdf}).  This allows us to define \emph{many} new distributions in terms of existing ones.

A \m{Bijector} implements how to transform a \m{Tensor} and how the input \m{Tensor}'s shape changes; this \m{Tensor} is presumed to be a random outcome possessing \\ \m{Distribution} shape semantics.
Three functions characterize how the \m{Tensor} is transformed:
\begin{enumerate}
\item
\m{forward} implements $x\mapsto F(x)$, and is used by \\ \m{TransformedDistribution.sample} to convert one random outcome into another. It also establishes the name of the bijector.
\item
\m{inverse} undoes the transformation $y\mapsto F^{-1}(y)$
and is used by \\\m{TransformedDistribution.log\_prob}.
\item
\m{inverse\_log\_det\_jacobian} computes \\ $\log |DF^{-1}(y)|$
and is used by \\ \m{TransformedDistribution.log\_prob} to adjust for how the volume changes by the transformation.
In certain settings, it is more numerically stable (or generally preferable) to implement the \m{forward\_log\_det\_jacobian}. Because forward and reverse $\log \circ \operatorname{det} \circ \operatorname{Jacobian}$s differ only in sign, either (or both) may be implemented.
\end{enumerate}

A \m{Bijector} also describes how it changes the \m{Tensor}'s shape so that \m{TransformedDistribution} can implement functions that compute event and batch shapes.
Most bijectors do not change the \m{Tensor}'s shape. Those which do implement \m{forward\_event\_shape\_tensor} and \\ \m{inverse\_event\_shape\_tensor}. Each takes an input shape (vector) and returns a new shape representing the \m{Tensor}'s event/batch shape after \m{forward} or \m{inverse} transformations. Excluding higher-order bijectors, currently only \m{SoftmaxCentered} changes the shape.\footnote{%
To implement $\operatorname{softmax}(x)=\exp(x)/\sum_i \exp(x_i)$ as a diffeomorophism, its \m{forward} appends a zero to the event and its \m{reverse} strips this padding. The result is a bijective map which avoids the fact that $\operatorname{softmax}(x)=\exp(x-c)/\sum_i \exp(x_i-c)$ for any $c$.}

Using a \m{Bijector}, \m{TransformedDistribution} automatically and efficiently implements \m{sample}, \m{log\_prob}, and \m{prob}. For bijectors with constant Jacobian such as  \m{Affine}, \m{TransformedDistribution} automatically implements statistics such as \m{mean}, \m{variance}, and \m{entropy}. The following example implements an affine-transformed Laplace distribution.
\begin{lstlisting}[basicstyle=\small\ttfamily]
vector_laplace = tfd.TransformedDistribution(
  distribution=tfd.Laplace(loc=0., scale=1.),
  bijector=tfb.Affine(
    shift=tf.Variable(tf.zeros(d)),
    scale_tril=tfd.fill_triangular(
      tf.Variable(tf.ones(d*(d+1)/2)))),
  event_shape=[d])
\end{lstlisting}
The distribution is learnable via \m{tf.Variable}s and that the \m{Affine} is parameterized by what is essentially the Cholesky of the covariance matrix. This makes the multivariate construction computationally efficient and more numerically stable; bijector caching (\Cref{sec:bijector_caching}) may even eliminate back substitution.

\subsection{Composability}
\label{sub:composability}

\m{Bijector}s can compose using higher-order \m{Bijector}s such as \m{Chain} and \m{Invert}. \Cref{fig:power_vae} illustrates a powerful example where the \m{arflow} method composes a sequence of autoregressive and permutation \m{Bijectors} to compactly describe an autoregressive flow \citep{kingma2016improved,papamakarios2017masked}.

The \m{Chain} bijector enables simple construction of rich \m{Distribution}s. Below we construct a multivariate logit-Normal with matrix-shaped events.
\begin{lstlisting}[basicstyle=\small\ttfamily]
matrix_logit_mvn =
  tfd.TransformedDistribution(
    distribution=tfd.Normal(0., 1.),
    bijector=tfb.Chain([
      tfb.Reshape([d, d]),
      tfb.SoftmaxCentered(),
      tfb.Affine(scale_diag=diag),
    ]),
    event_shape=[d * d])
\end{lstlisting}
The \m{Invert} bijector effectively doubles the number of bijectors by swapping \m{forward} and \m{inverse}. It is useful in situations such as the Gumbel construction in \Cref{sub:higher}. It is also useful for transforming constrained continuous distributions onto an unconstrained real-valued space. For example:
\begin{lstlisting}[basicstyle=\small\ttfamily]
softminus_gamma = tfd.TransformedDistribution(
 distribution=tfd.Gamma(
   concentration=alpha,
   rate=beta),
 bijector=tfb.Invert(tfb.Softplus()))
\end{lstlisting}
This performs a softplus-inversion to robustly transform \m{Gamma} to be unconstrained. This enables a key component of automated algorithms such as automatic differentiation variational inference \citep{kucukelbir2017automatic} and the No U-Turn Sampler \cite{hoffman2014nuts}. They only operate on real-valued spaces, so unconstraints expand their scope.

\subsection{Caching}
\label{sec:bijector_caching}

\m{Bijector}s automatically cache input/output pairs of operations,
including the $\log \circ \operatorname{det} \circ
\operatorname{Jacobian}$. This is advantageous when the \m{inverse}
calculation is slow, numerically unstable, or not easily
implementable. A cache hit occurs when computing the probability of
results of \m{sample}. That is, if $q(x)$ is the distribution
associated with $x=f(\varepsilon)$ and $\varepsilon\sim r$, then
caching lowers the cost of computing $q(x_i)$ since \\ $q(x_i) = r((f^{-1} \circ f)(\varepsilon_i)) \left|\left(\frac{\partial}{\partial \varepsilon} \circ f^{-1} \circ f\right)(\varepsilon_i)\right|^{-1} = r(\varepsilon_i)$.

Because TensorFlow follows a deferred execution model, \m{Bijector} caching is nominal; it has zero memory or computational cost. The \m{Bijector} base class merely replaces one graph node for another already existing node. Since the existing node (``cache hit'') is already an execution dependency, the only cost of ``caching'' is during graph construction.

Caching is computationally and numerically beneficial for importance sampling algorithms, which compute expectations. They weight by a drawn samples' reciprocal probability. Namely,
\begin{align*}
\mu &= \int f(x)p(x)dx \\
&=\int\frac{f(x)p(x)}{q(x)}q(x)dx\\
&= \lim_{n\to \infty} n^{-1}\sum_i^n\frac{f(x_i)p(x_i)}{q(x_i)}, \text{ where } x_i \iid q.
\end{align*}
Caching also nicely complements black-box variational inference
algorithms \citep{ranganath2014black,kucukelbir2017automatic}. In
these procedures, the approximate posterior distribution only computes
\\ \m{log\_prob} over its own \m{sample}s. In this setting the sample's preimage ($\varepsilon_i$) is known without computing $f^{-1}(x_i)$.

\m{MultivariateNormalTriL} is implemented as a \\ \m{TransformedDistribution} with the \m{Affine} bijector. Caching removes the need for quadratic complexity back substitution. For an InverseAutoregressiveFlow \citep{kingma2016improved},
\begin{lstlisting}[basicstyle=\small\ttfamily]
laplace_iaf = tfd.TransformedDistribution(
  distribution=tfd.Laplace(loc=0., scale=1.),
  bijector=tfb.Invert(
    tfb.MaskedAutoregressiveFlow(
      shift_and_log_scale_fn=tfb.\
      masked_autoregressive_default_template(
          hidden_layers))),
  event_shape=[d])
\end{lstlisting}
caching reduces the overall complexity from quadratic to linear (in event size).

\subsection{Smooth Coverings}
\label{subsection:smooth-covering}
The \m{Bijector} framework extends to non-injective transformations, i.e., \emph{smooth coverings} \citep{spivak2010comprehensive}.\footnote{%
\m{Bijector} caching is currently not supported for smooth coverings.}
Formally, a smooth covering is a continuous function $F$ on the entire domain $D$ where, ignoring sets of measure zero, the domain can be partitioned as a finite union $D = D_1\cup\cdots\cup D_K$ such that each restriction $F:D_k\to F(D)$ is a diffeomorphism. Examples include \m{AbsValue} and \m{Square}. We implement them by having the \m{inverse} method return the set inverse $F^{-1}(y) := \{x \;:\; F(x) = y\}$ as a tuple.

Smooth covering \m{Bijector}s let us easily build half-distributions, which allocate probability mass over only the positive half-plane of the original distribution. For example, we build a half-Cauchy distribution as
\begin{lstlisting}[basicstyle=\small\ttfamily]
half_cauchy = tfd.TransformedDistribution(
  bijector=tfb.AbsValue(),
  distribution=tfd.Cauchy(loc=0., scale=1.))
\end{lstlisting}
The half-Cauchy and half-Student t distributions are often used as ``weakly informative'' priors, which exhibit heavy tails, for variance parameters in hierarchical models \citep{gelman2006prior}.

\section{Applications}
\label{sec:applications}

We described two abstractions: \m{Distribution}s and \\ \m{Bijector}s. Recall \Cref{fig:generic_vae,fig:simple_vae,fig:power_vae}, where we showed the power of combining these abstractions for changing from simple to state-of-the-art variational auto-encoders. Below we show additional applications of TensorFlow Distributions as part of the TensorFlow ecosystem.

\subsection{Kernel Density Estimators}

A kernel density estimator (KDE) is a nonparametric estimator of an unknown probability distribution \citep{wasserman2013all}. Kernel density estimation provides a fundamental smoothing operation on finite samples that is useful across many applications. With TensorFlow Distributions,
KDEs can be flexibly constructed as a \m{MixtureSameFamily}.
Given a finite set of points \m{x} in $\mathbb{R}^D$, we write

\begin{lstlisting}
f = lambda x: tfd.Independent(tfd.Normal(
  loc=x, scale=1.))
n = x.shape[0].value
kde = tfd.MixtureSameFamily(
  mixture_distribution=tfd.Categorical(
    probs=[1 / n] * n),
  components_distribution=f(x))
\end{lstlisting}

Here, \m{f} is a callable taking \m{x} as input and returns a distribution. Above, we use an independent $D$-dimensional \m{Normal} distribution (equivalent to \\ \m{MultivariateNormalDiag}), which induces a Gaussian kernel density estimator.


Changing the callable extends the KDE to alternative distribution-based kernels. For example, we can use
\m{lambda x: MultivariateNormalTriL(loc=x)} for a multivariate kernel, and alternative distributions such as \m{lambda x: Independent(StudentT(loc=x, scale=0.5, df=3)}.
The same concept also applies for bootstrap techniques \citep{efron1994introduction}. We can employ parametric bootstrap or stratified sampling to replace the equal mixing weights.

\subsection{PixelCNN Distribution}
\label{sub:pixelcnn}

Building from the KDE example above, we now show a modern, high-dimensional density estimator. \Cref{fig:pixelcnn} builds a PixelCNN \citep{vandenoord2016pixel} as a fully-visible autoregressive distribution on \m{images}, which is a batch of $32\times 32\times 3$ RGB pixel images from Small ImageNet.

The variable \m{x} is the pixelCNN distribution. It makes use of the higher-order \m{Autoregressive} distribution, which takes as input a Python callable and number of autoregressive steps. The Python callable takes in currently observed features and returns the per-time step distribution.
\m{-tf.reduce\_sum(x.log\_prob(images))} is the loss function for maximum likelihood training; \m{x.sample} generates new images.

We also emphasize modularity. Note here, we used the pixelCNN as a fully-visible distribution. This differs from \Cref{fig:power_vae} which employs pixelCNN as a decoder (conditional likelihood).


\begin{figure}[tb]
\begin{lstlisting}[language=python]
import pixelcnn

def pixelcnn_dist(params, x_shape=(32, 32, 3)):
  def _logit_func(features):
    # implement single autoregressive step
    # on observed features
    logits = pixelcnn(features)
    return logits
  logit_template = tf.make_template(
    "pixelcnn", _logit_func)
  make_dist = lambda x: tfd.Independent(
    tfd.Bernoulli(logit_template(x)))
  return tfd.Autoregressive(
    make_dist, tf.reduce_prod(x_shape)))

x = pixelcnn_dist()
loss = -tf.reduce_sum(x.log_prob(images))
train = tf.train.AdamOptimizer(
    ).minimize(loss)  # run for training
generate = x.sample()  # run for generation
\end{lstlisting}
\vspace{-2ex}
\caption{PixelCNN distribution on images. It uses \m{Autoregressive}, which takes as input a callable returning a distribution per time step.}
\label{fig:pixelcnn}
\vspace{-2ex}
\end{figure}







\subsection{Edward Probabilistic Programs}
\label{sub:edward}

We describe how TensorFlow Distributions enables Edward as a backend. In particular, note that non-goals in TensorFlow Distributions can be accomplished at higher-level abstractions. Here, Edward wraps TensorFlow Distributions as random variables, associating each \m{Distribution} to a random outcome \m{Tensor} (calling \m{sample}) in the TensorFlow graph. This enables a calculus where TensorFlow ops can be applied directly to Edward random variables; this is a non-goal for TensorFlow Distributions.

\begin{figure}[tb]
\begin{lstlisting}[language=python]
from edward.models import Normal

z = x = []
z[0] = Normal(loc=tf.zeros(K),scale=tf.ones(K))
h = tf.layers.dense(
  z[0], 512, activation=tf.nn.relu)
loc = tf.layers.dense(h, D, activation=None)
x[0] = Normal(loc=loc, scale=0.5)
for t in range(1, T):
  inputs = tf.concat([z[t - 1], x[t - 1]], 0)
  loc = tf.layers.dense(
    inputs, K, activation=tf.tanh)
  z[t] = Normal(loc=loc, scale=0.1)
  h = tf.layers.dense(
    z[t], 512, activation=tf.nn.relu)
  loc = tf.layers.dense(h, D, activation=None)
  x[t] = Normal(loc=loc, scale=0.5)
\end{lstlisting}
\vspace{-2ex}
\caption{Stochastic recurrent neural network, which is
a state space model with nonlinear dynamics.
The transition mimicks a recurrent tanh cell and the
omission is multi-layered.
}
\label{fig:stochastic-rnn}
\end{figure}

As an example, \Cref{fig:stochastic-rnn} implements
a stochastic recurrent neural network (RNN), which is an RNN whose
hidden state is random \citep{bayer2014learning}.
Formally, for $T$ time steps, the model specifies the joint distribution
\begin{align*}
p(\mathbf{x}, \mathbf{z}) &=
\operatorname{Normal}(\mathbf{z}_1\mid \mathbf{0}, \mathbf{I})
\prod_{t=2}^{T}
\operatorname{Normal}(\mathbf{z}_t\mid f(\mathbf{z}_{t-1}), 0.1) \\
&\quad
\prod_{t=1}^{T}
\operatorname{Normal}(\mathbf{x}_t\mid g(\mathbf{z}_{t}), 0.5)
,
\end{align*}
where each time step in an observed real-valued sequence
$\mathbf{x}=[\mathbf{x}_1,\ldots,\mathbf{x}_T]\in\mathbb{R}^{T\times
D}$ is associated with an
unobserved state $\mathbf{z}_t\in\mathbb{R}^K$; the initial latent
variable $\mathbf{z}_1$ is drawn randomly from a standard normal.
The noise standard deviations
are fixed and broadcasted over the batch.
The latent variable and likelihood are parameterized by neural networks.

The program is generative: starting from a latent state, it unrolls state dynamics through time. Given this program and data, Edward's algorithms enable approximate inference (a second non-goal of TensorFlow Distributions).

\subsection{TensorFlow Estimator API}
\label{sub:estimator}

As part of the TensorFlow ecosystem, TensorFlow Distributions complements other TensorFlow libraries. We show how it complements TensorFlow Estimator.

\Cref{fig:estimator} demonstrates multivariate linear regression in the presence of heteroscedastic noise,
\begin{align*}
U &\sim \operatorname{\sf MultivariateNormal}(0, I_d) \\
Y &= \Sigma^{\frac{1}{2}}(X) U + \mu(X)
\end{align*}
where $\Sigma:\mathbb{R}^d \to \{Z \in \mathbb{R}^{d\times d} : Z \succcurlyeq 0\}$, $\mu:\mathbb{R}^d \to \mathbb{R}^d$, and $\Sigma^{\frac{1}{2}}$ denotes the Cholesky decomposition. Adding more \m{tf.layers} to the parameterization of the \\ \m{MultivariateNormalTriL} enables learning nonlinear transformations. ($\Sigma=I_d$ would be appropriate in homoscedastic regimes.)

\begin{figure}[tb]
\begin{lstlisting}[language=python]
def mvn_regression_fn(
    features, labels, mode, params=None):
  d = features.shape[-1].value
  mvn = tfd.MultivariateNormalTriL(
    loc=tf.layers.dense(features, d),
    scale_tril=tfd.fill_triangular(
      tf.layers.dense(features, d*(d+1)/2)))
  if mode == tf.estimator.ModeKeys.PREDICT:
    return mvn.mean()
  loss = -tf.reduce_sum(mvn.log_prob(labels))
  if mode == tf.estimator.ModeKeys.EVAL:
    metric_fn = lambda x,y:
      tf.metrics.mean_squared_error(x, y)
    return tpu_estimator.TPUEstimatorSpec(
      mode=mode,
      loss=loss,
      eval_metrics=(
        metric_fn, [labels, mvn.mean()]))
  optimizer = tf.train.AdamOptimizer()
  if FLAGS.use_tpu:
    optimizer = tpu_optimizer.CrossShardOptimizer(optimizer)
  train_op = optimizer.minimize(loss)
  return tpu_estimator.TPUEstimatorSpec(mode=mode, loss=loss, train_op=train_op)

# TPUEstimator Boilerplate.
run_config = tpu_config.RunConfig(
  master=FLAGS.master,
  model_dir=FLAGS.model_dir,
  session_config=tf.ConfigProto(
    allow_soft_placement=True,
    log_device_placement=True),
  tpu_config=tpu_config.TPUConfig(
    FLAGS.iterations,
    FLAGS.num_shards))
estimator = tpu_estimator.TPUEstimator(
  model_fn=mvn_regression_fn,
  config=run_config,
  use_tpu=FLAGS.use_tpu,
  train_batch_size=FLAGS.batch_size,
  eval_batch_size=FLAGS.batch_size)
\end{lstlisting}
\vspace{-2ex}
\caption{Multivariate regression with TPUs.}
\label{fig:estimator}
\vspace{-2ex}
\end{figure}

Using \m{Distribution}s to build \m{Estimator}s is natural and ergonomic. We use \m{TPUEstimator} in particular, which extends \m{Estimator} with configurations for TPUs \citep{jouppi2017indatacenter}. Together, \m{Distribution}s and \m{Estimator}s provide a simple, scalable platform for efficiently deploying training, evaluation, and prediction over diverse hardware and network topologies.


\Cref{fig:estimator} only writes the \texttt{Estimator} object.
For training, call \m{estimator.train()};
for evaluation, call \\ \m{estimator.evaluate()}; for prediction, call \\ \m{estimator.predict()}. Each takes an input function to load in data.


%
%
%
%
%

\section{Discussion}

The TensorFlow Distributions library implements a vision of probability theory adapted to the modern deep-learning paradigm of end-to-end differentiable computation.
\m{Distribution}s provides a collection of
56
distributions with fast, numerically stable methods for sampling,
computing log densities, and many statistics. \\ \m{Bijector}s provides a collection of
22
composable transformations with efficient volume-tracking and caching.

Although Tensorflow Distributions is relatively new, it has already seen widespread adoption both inside and outside of Google. External developers have built on it to design probabilistic programming and statistical systems including Edward \citep{tran2016edward} and Greta \citep{golding2017greta}.
Further, \m{Distribution} and \m{Bijector} is being used as the design basis for similar functionality in the PyTorch computational graph framework \citep{pytorch2017pytorch}, as well as the Pyro and ZhuSuan probabilistic programming systems \citep{pyro2017pyro, shi2017zhusuan}.

Looking forward, we plan to continue expanding the set of supported \m{Distributions} and \m{Bijectors}.
We intend to expand the \m{Distribution} interface to include supports, e.g., real-valued, positive, unit interval, etc., as a class property.
%
We are also exploring the possibility of exposing exponential family
structure, for example providing separate \m{unnormalized\_log\_prob}
and \\ \m{log\_normalizer} methods where appropriate.

As part of the trend towards hardware-accelerated linear algebra, we are working to ensure that all distribution and bijector methods are compatible with TPUs \citep{jouppi2017indatacenter}, including special functions such as gamma, as well as rejection sampling-based (e.g., Gamma) and while-loop based sampling mechanisms (e.g., Poisson). We also aim to natively support Distributions over \m{SparseTensor}s.



\begin{acks}
  We thank Jasper Snoek for feedback and comments, and Kevin Murphy for thoughtful discussions since the beginning of TensorFlow Distributions.
  DT is supported by a Google Ph.D. Fellowship in Machine Learning and an Adobe Research Fellowship.
\end{acks}

\bibliographystyle{ACM-Reference-Format}
\bibliography{bib}


\begin{thebibliography}{51}


\ifx \showCODEN    \undefined \def \showCODEN     #1{\unskip}     \fi
\ifx \showDOI      \undefined \def \showDOI       #1{#1}\fi
\ifx \showISBNx    \undefined \def \showISBNx     #1{\unskip}     \fi
\ifx \showISBNxiii \undefined \def \showISBNxiii  #1{\unskip}     \fi
\ifx \showISSN     \undefined \def \showISSN      #1{\unskip}     \fi
\ifx \showLCCN     \undefined \def \showLCCN      #1{\unskip}     \fi
\ifx \shownote     \undefined \def \shownote      #1{#1}          \fi
\ifx \showarticletitle \undefined \def \showarticletitle #1{#1}   \fi
\ifx \showURL      \undefined \def \showURL       {\relax}        \fi
\providecommand\bibfield[2]{#2}
\providecommand\bibinfo[2]{#2}
\providecommand\natexlab[1]{#1}
\providecommand\showeprint[2][]{arXiv:#2}

\bibitem[\protect\citeauthoryear{Abadi, Agarwal, Barham, Brevdo, Chen, Citro,
  Corrado, Davis, Dean, Devin, Ghemawat, Goodfellow, Harp, Irving, Isard, Jia,
  Jozefowicz, Kaiser, Kudlur, Levenberg, Man\'{e}, Monga, Moore, Murray, Olah,
  Schuster, Shlens, Steiner, Sutskever, Talwar, Tucker, Vanhoucke, Vasudevan,
  Vi\'{e}gas, Vinyals, Warden, Wattenberg, Wicke, Yu, and Zheng}{Abadi
  et~al\mbox{.}}{2015}]%
        {abadi2016tensorflow}
\bibfield{author}{\bibinfo{person}{Mart\'{\i}n Abadi}, \bibinfo{person}{Ashish
  Agarwal}, \bibinfo{person}{Paul Barham}, \bibinfo{person}{Eugene Brevdo},
  \bibinfo{person}{Zhifeng Chen}, \bibinfo{person}{Craig Citro},
  \bibinfo{person}{Greg~S. Corrado}, \bibinfo{person}{Andy Davis},
  \bibinfo{person}{Jeffrey Dean}, \bibinfo{person}{Matthieu Devin},
  \bibinfo{person}{Sanjay Ghemawat}, \bibinfo{person}{Ian Goodfellow},
  \bibinfo{person}{Andrew Harp}, \bibinfo{person}{Geoffrey Irving},
  \bibinfo{person}{Michael Isard}, \bibinfo{person}{Yangqing Jia},
  \bibinfo{person}{Rafal Jozefowicz}, \bibinfo{person}{Lukasz Kaiser},
  \bibinfo{person}{Manjunath Kudlur}, \bibinfo{person}{Josh Levenberg},
  \bibinfo{person}{Dan Man\'{e}}, \bibinfo{person}{Rajat Monga},
  \bibinfo{person}{Sherry Moore}, \bibinfo{person}{Derek Murray},
  \bibinfo{person}{Chris Olah}, \bibinfo{person}{Mike Schuster},
  \bibinfo{person}{Jonathon Shlens}, \bibinfo{person}{Benoit Steiner},
  \bibinfo{person}{Ilya Sutskever}, \bibinfo{person}{Kunal Talwar},
  \bibinfo{person}{Paul Tucker}, \bibinfo{person}{Vincent Vanhoucke},
  \bibinfo{person}{Vijay Vasudevan}, \bibinfo{person}{Fernanda Vi\'{e}gas},
  \bibinfo{person}{Oriol Vinyals}, \bibinfo{person}{Pete Warden},
  \bibinfo{person}{Martin Wattenberg}, \bibinfo{person}{Martin Wicke},
  \bibinfo{person}{Yuan Yu}, {and} \bibinfo{person}{Xiaoqiang Zheng}.}
  \bibinfo{year}{2015}\natexlab{}.
\newblock \bibinfo{title}{{TensorFlow}: Large-Scale Machine Learning on
  Heterogeneous Systems}.
\newblock   (\bibinfo{year}{2015}).
\newblock
\showURL{%
\url{https://www.tensorflow.org/}}
\newblock
\shownote{Software available from tensorflow.org.}


\bibitem[\protect\citeauthoryear{Ackerman, Freer, and Roy}{Ackerman
  et~al\mbox{.}}{2016}]%
        {ackerman2016exchangeable}
\bibfield{author}{\bibinfo{person}{Nathanel~L Ackerman},
  \bibinfo{person}{Cameron~E Freer}, {and} \bibinfo{person}{Daniel~M Roy}.}
  \bibinfo{year}{2016}\natexlab{}.
\newblock \showarticletitle{Exchangeable random primitives}. In
  \bibinfo{booktitle}{{\em Workshop on Probabilistic Programming Semantics}}.
  \bibinfo{pages}{2016}.
\newblock


\bibitem[\protect\citeauthoryear{Aldous}{Aldous}{1985}]%
        {aldous1985exchangeability}
\bibfield{author}{\bibinfo{person}{David~J Aldous}.}
  \bibinfo{year}{1985}\natexlab{}.
\newblock \showarticletitle{Exchangeability and related topics}.
\newblock In \bibinfo{booktitle}{{\em {\'E}cole d'{\'E}t{\'e} de
  Probabilit{\'e}s de Saint-Flour XIII—1983}}. \bibinfo{publisher}{Springer},
  \bibinfo{pages}{1--198}.
\newblock


\bibitem[\protect\citeauthoryear{Anonymous}{Anonymous}{2017}]%
        {anonymous2017generative}
\bibfield{author}{\bibinfo{person}{Anonymous}.}
  \bibinfo{year}{2017}\natexlab{}.
\newblock \showarticletitle{Generative Models for Data Efficiency and Alignment
  in Language}.
\newblock \bibinfo{journal}{{\em OpenReview\/}} (\bibinfo{year}{2017}).
\newblock


\bibitem[\protect\citeauthoryear{Bayer and Osendorfer}{Bayer and
  Osendorfer}{2014}]%
        {bayer2014learning}
\bibfield{author}{\bibinfo{person}{Justin Bayer} {and}
  \bibinfo{person}{Christian Osendorfer}.} \bibinfo{year}{2014}\natexlab{}.
\newblock \showarticletitle{{Learning Stochastic Recurrent Networks}}.
\newblock \bibinfo{journal}{{\em arXiv.org\/}} (\bibinfo{year}{2014}).
\newblock
\showeprint{1411.7610v3}


\bibitem[\protect\citeauthoryear{Becker and Chambers}{Becker and
  Chambers}{1984}]%
        {becker1984s}
\bibfield{author}{\bibinfo{person}{Richard~A Becker} {and}
  \bibinfo{person}{John~M Chambers}.} \bibinfo{year}{1984}\natexlab{}.
\newblock \bibinfo{booktitle}{{\em S: an interactive environment for data
  analysis and graphics}}.
\newblock \bibinfo{publisher}{CRC Press}.
\newblock


\bibitem[\protect\citeauthoryear{Blei and Lafferty}{Blei and Lafferty}{2006}]%
        {blei2006correlated}
\bibfield{author}{\bibinfo{person}{David~M Blei} {and} \bibinfo{person}{John
  Lafferty}.} \bibinfo{year}{2006}\natexlab{}.
\newblock \showarticletitle{{Correlated topic models}}. In
  \bibinfo{booktitle}{{\em Neural Information Processing Systems}}.
\newblock


\bibitem[\protect\citeauthoryear{Box and Muller}{Box and Muller}{1958}]%
        {boxmullertransform}
\bibfield{author}{\bibinfo{person}{G.~E.~P. Box} {and}
  \bibinfo{person}{Mervin~E. Muller}.} \bibinfo{year}{1958}\natexlab{}.
\newblock \showarticletitle{{A Note on the Generation of Random Normal
  Deviates}}.
\newblock \bibinfo{journal}{{\em The Annals of Mathematical Statistics\/}}
  (\bibinfo{year}{1958}), \bibinfo{pages}{610--611}.
\newblock


\bibitem[\protect\citeauthoryear{Carpenter, Gelman, Hoffman, Lee, Goodrich,
  Betancourt, Brubaker, Guo, Li, and Riddell}{Carpenter et~al\mbox{.}}{2016}]%
        {carpenter2016stan}
\bibfield{author}{\bibinfo{person}{Bob Carpenter}, \bibinfo{person}{Andrew
  Gelman}, \bibinfo{person}{Matthew~D Hoffman}, \bibinfo{person}{Daniel Lee},
  \bibinfo{person}{Ben Goodrich}, \bibinfo{person}{Michael Betancourt},
  \bibinfo{person}{Marcus Brubaker}, \bibinfo{person}{Jiqiang Guo},
  \bibinfo{person}{Peter Li}, {and} \bibinfo{person}{Allen Riddell}.}
  \bibinfo{year}{2016}\natexlab{}.
\newblock \showarticletitle{{Stan: a probabilistic programming language}}.
\newblock \bibinfo{journal}{{\em Journal of Statistical Software\/}}
  (\bibinfo{year}{2016}).
\newblock


\bibitem[\protect\citeauthoryear{Carpenter, Hoffman, Brubaker, Lee, Li, and
  Betancourt}{Carpenter et~al\mbox{.}}{2015}]%
        {carpenter2015stan}
\bibfield{author}{\bibinfo{person}{Bob Carpenter}, \bibinfo{person}{Matthew~D
  Hoffman}, \bibinfo{person}{Marcus Brubaker}, \bibinfo{person}{Daniel Lee},
  \bibinfo{person}{Peter Li}, {and} \bibinfo{person}{Michael Betancourt}.}
  \bibinfo{year}{2015}\natexlab{}.
\newblock \showarticletitle{The {Stan} math library: Reverse-mode automatic
  differentiation in C++}.
\newblock \bibinfo{journal}{{\em arXiv preprint arXiv:1509.07164\/}}
  (\bibinfo{year}{2015}).
\newblock


\bibitem[\protect\citeauthoryear{Cover and Thomas}{Cover and Thomas}{1991}]%
        {cover1991elements}
\bibfield{author}{\bibinfo{person}{Thomas~M Cover} {and} \bibinfo{person}{Joy~A
  Thomas}.} \bibinfo{year}{1991}\natexlab{}.
\newblock \bibinfo{booktitle}{{\em {Elements of Information Theory}}}.
\newblock \bibinfo{publisher}{Wiley Series in Telecommunications and Signal
  Processing}.
\newblock


\bibitem[\protect\citeauthoryear{Csisz{\'a}r}{Csisz{\'a}r}{1963}]%
        {csiszar1963informationstheoretische}
\bibfield{author}{\bibinfo{person}{Imre Csisz{\'a}r}.}
  \bibinfo{year}{1963}\natexlab{}.
\newblock \showarticletitle{Eine informationstheoretische Ungleichung und ihre
  Anwendung auf Beweis der Ergodizitaet von Markoffschen Ketten}.
\newblock \bibinfo{journal}{{\em Magyer Tud. Akad. Mat. Kutato Int. Koezl.\/}}
  \bibinfo{volume}{8} (\bibinfo{year}{1963}), \bibinfo{pages}{85--108}.
\newblock


\bibitem[\protect\citeauthoryear{Efron and Tibshirani}{Efron and
  Tibshirani}{1994}]%
        {efron1994introduction}
\bibfield{author}{\bibinfo{person}{Bradley Efron} {and}
  \bibinfo{person}{Robert~J Tibshirani}.} \bibinfo{year}{1994}\natexlab{}.
\newblock \bibinfo{booktitle}{{\em An introduction to the bootstrap}}.
\newblock \bibinfo{publisher}{CRC press}.
\newblock


\bibitem[\protect\citeauthoryear{Fu}{Fu}{2006}]%
        {fu-gradient-estimation}
\bibfield{author}{\bibinfo{person}{M.C.\ Fu}.} \bibinfo{year}{2006}\natexlab{}.
\newblock \bibinfo{booktitle}{{\em Simulation}}. \bibinfo{series}{Handbook in
  Operations Research and Management Science}, Vol.~\bibinfo{volume}{13}.
\newblock \bibinfo{publisher}{North Holland}.
\newblock
\showISBNx{9780444514288}


\bibitem[\protect\citeauthoryear{Garcia-Garcia and Williamson}{Garcia-Garcia
  and Williamson}{2012}]%
        {garcia2012divergences}
\bibfield{author}{\bibinfo{person}{Dario Garcia-Garcia} {and}
  \bibinfo{person}{Robert~C Williamson}.} \bibinfo{year}{2012}\natexlab{}.
\newblock \showarticletitle{Divergences and risks for multiclass experiments}.
  In \bibinfo{booktitle}{{\em Conference on Learning Theory}}.
\newblock


\bibitem[\protect\citeauthoryear{Gelman et~al\mbox{.}}{Gelman
  et~al\mbox{.}}{2006}]%
        {gelman2006prior}
\bibfield{author}{\bibinfo{person}{Andrew Gelman} {et~al\mbox{.}}}
  \bibinfo{year}{2006}\natexlab{}.
\newblock \showarticletitle{Prior distributions for variance parameters in
  hierarchical models (comment on article by Browne and Draper)}.
\newblock \bibinfo{journal}{{\em Bayesian analysis\/}} \bibinfo{volume}{1},
  \bibinfo{number}{3} (\bibinfo{year}{2006}), \bibinfo{pages}{515--534}.
\newblock


\bibitem[\protect\citeauthoryear{Golding}{Golding}{2017}]%
        {golding2017greta}
\bibfield{author}{\bibinfo{person}{Nick Golding}.}
  \bibinfo{year}{2017}\natexlab{}.
\newblock \bibinfo{booktitle}{{\em greta: Simple and Scalable Statistical
  Modelling in R}}.
\newblock
\showURL{%
\url{https://CRAN.R-project.org/package=greta}}
\newblock
\shownote{R package version 0.2.0.}


\bibitem[\protect\citeauthoryear{G{\'o}mez-Bombarelli, Duvenaud,
  Hern{\'a}ndez-Lobato, Aguilera-Iparraguirre, Hirzel, Adams, and
  Aspuru-Guzik}{G{\'o}mez-Bombarelli et~al\mbox{.}}{2016}]%
        {gomez2016automatic}
\bibfield{author}{\bibinfo{person}{Rafael G{\'o}mez-Bombarelli},
  \bibinfo{person}{David Duvenaud}, \bibinfo{person}{Jos{\'e}~Miguel
  Hern{\'a}ndez-Lobato}, \bibinfo{person}{Jorge Aguilera-Iparraguirre},
  \bibinfo{person}{Timothy~D Hirzel}, \bibinfo{person}{Ryan~P Adams}, {and}
  \bibinfo{person}{Al{\'a}n Aspuru-Guzik}.} \bibinfo{year}{2016}\natexlab{}.
\newblock \showarticletitle{Automatic chemical design using a data-driven
  continuous representation of molecules}.
\newblock \bibinfo{journal}{{\em arXiv preprint arXiv:1610.02415\/}}
  (\bibinfo{year}{2016}).
\newblock


\bibitem[\protect\citeauthoryear{Goodfellow, Pouget-Abadie, Mirza, Xu,
  Warde-Farley, Ozair, Courville, and Bengio}{Goodfellow et~al\mbox{.}}{2014}]%
        {goodfellow2014generative}
\bibfield{author}{\bibinfo{person}{Ian Goodfellow}, \bibinfo{person}{Jean
  Pouget-Abadie}, \bibinfo{person}{M Mirza}, \bibinfo{person}{Bing Xu},
  \bibinfo{person}{David Warde-Farley}, \bibinfo{person}{Sherjil Ozair},
  \bibinfo{person}{Aaron Courville}, {and} \bibinfo{person}{Yoshua Bengio}.}
  \bibinfo{year}{2014}\natexlab{}.
\newblock \showarticletitle{{Generative Adversarial Nets}}. In
  \bibinfo{booktitle}{{\em Neural Information Processing Systems}}.
\newblock


\bibitem[\protect\citeauthoryear{Hoffman and Gelman}{Hoffman and
  Gelman}{2014}]%
        {hoffman2014nuts}
\bibfield{author}{\bibinfo{person}{Matthew~D Hoffman} {and}
  \bibinfo{person}{Andrew Gelman}.} \bibinfo{year}{2014}\natexlab{}.
\newblock \showarticletitle{{The no-U-turn sampler: adaptively setting path
  lengths in Hamiltonian Monte Carlo}}.
\newblock \bibinfo{journal}{{\em Journal of Machine Learning Research\/}}
  \bibinfo{volume}{15} (\bibinfo{year}{2014}), \bibinfo{pages}{1593--1623}.
\newblock


\bibitem[\protect\citeauthoryear{Ihaka and Gentleman}{Ihaka and
  Gentleman}{1996}]%
        {ihaka1996r}
\bibfield{author}{\bibinfo{person}{Ross Ihaka} {and} \bibinfo{person}{Robert
  Gentleman}.} \bibinfo{year}{1996}\natexlab{}.
\newblock \showarticletitle{R: A Language for Data Analysis and Graphics}.
\newblock \bibinfo{journal}{{\em Journal of Computational and Graphical
  Statistics\/}} \bibinfo{volume}{5}, \bibinfo{number}{3}
  (\bibinfo{year}{1996}), \bibinfo{pages}{299--314}.
\newblock
\showDOI{%
\url{https://doi.org/10.1080/10618600.1996.10474713}}


\bibitem[\protect\citeauthoryear{Jang, Gu, and Poole}{Jang
  et~al\mbox{.}}{2017}]%
        {jang2017categorical}
\bibfield{author}{\bibinfo{person}{Eric Jang}, \bibinfo{person}{Shixiang Gu},
  {and} \bibinfo{person}{Ben Poole}.} \bibinfo{year}{2017}\natexlab{}.
\newblock \showarticletitle{Categorical reparameterization with
  gumbel-softmax}. In \bibinfo{booktitle}{{\em International Conference on
  Learning Representations}}.
\newblock


\bibitem[\protect\citeauthoryear{Johnson, Duvenaud, Wiltschko, Adams, and
  Datta}{Johnson et~al\mbox{.}}{2016}]%
        {johnson2016composing}
\bibfield{author}{\bibinfo{person}{Matthew Johnson}, \bibinfo{person}{David~K
  Duvenaud}, \bibinfo{person}{Alex Wiltschko}, \bibinfo{person}{Ryan~P Adams},
  {and} \bibinfo{person}{Sandeep~R Datta}.} \bibinfo{year}{2016}\natexlab{}.
\newblock \showarticletitle{{Composing graphical models with neural networks
  for structured representations and fast inference}}. In
  \bibinfo{booktitle}{{\em Neural Information Processing Systems}}.
\newblock


\bibitem[\protect\citeauthoryear{Jones, Oliphant, Peterson,
  et~al\mbox{.}}{Jones et~al\mbox{.}}{2001}]%
        {jones2001scipy}
\bibfield{author}{\bibinfo{person}{Eric Jones}, \bibinfo{person}{Travis
  Oliphant}, \bibinfo{person}{Pearu Peterson}, {et~al\mbox{.}}}
  \bibinfo{year}{2001}\natexlab{}.
\newblock \bibinfo{title}{{SciPy}: Open source scientific tools for {Python}}.
\newblock   (\bibinfo{year}{2001}).
\newblock
\showURL{%
\url{http://www.scipy.org/}}


\bibitem[\protect\citeauthoryear{Jouppi, Young, Patil, Patterson, Agrawal,
  Bajwa, Bates, Bhatia, Boden, Borchers, Boyle, Cantin, Chao, Clark, Coriell,
  Daley, Dau, Dean, Gelb, Ghaemmaghami, Gottipati, Gulland, Hagmann, Ho,
  Hogberg, Hu, Hundt, Hurt, Ibarz, Jaffey, Jaworski, Kaplan, Khaitan, Koch,
  Kumar, Lacy, Laudon, Law, Le, Leary, Liu, Lucke, Lundin, MacKean, Maggiore,
  Mahony, Miller, Nagarajan, Narayanaswami, Ni, Nix, Norrie, Omernick,
  Penukonda, Phelps, Ross, Ross, Salek, Samadiani, Severn, Sizikov, Snelham,
  Souter, Steinberg, Swing, Tan, Thorson, Tian, Toma, Tuttle, Vasudevan,
  Walter, Wang, Wilcox, and Yoon}{Jouppi et~al\mbox{.}}{2017}]%
        {jouppi2017indatacenter}
\bibfield{author}{\bibinfo{person}{Norman~P Jouppi}, \bibinfo{person}{Cliff
  Young}, \bibinfo{person}{Nishant Patil}, \bibinfo{person}{David Patterson},
  \bibinfo{person}{Gaurav Agrawal}, \bibinfo{person}{Raminder Bajwa},
  \bibinfo{person}{Sarah Bates}, \bibinfo{person}{Suresh Bhatia},
  \bibinfo{person}{Nan Boden}, \bibinfo{person}{Al Borchers},
  \bibinfo{person}{Rick Boyle}, \bibinfo{person}{Pierre-luc Cantin},
  \bibinfo{person}{Clifford Chao}, \bibinfo{person}{Chris Clark},
  \bibinfo{person}{Jeremy Coriell}, \bibinfo{person}{Mike Daley},
  \bibinfo{person}{Matt Dau}, \bibinfo{person}{Jeffrey Dean},
  \bibinfo{person}{Ben Gelb}, \bibinfo{person}{Tara~Vazir Ghaemmaghami},
  \bibinfo{person}{Rajendra Gottipati}, \bibinfo{person}{William Gulland},
  \bibinfo{person}{Robert Hagmann}, \bibinfo{person}{C~Richard Ho},
  \bibinfo{person}{Doug Hogberg}, \bibinfo{person}{John Hu},
  \bibinfo{person}{Robert Hundt}, \bibinfo{person}{Dan Hurt},
  \bibinfo{person}{Julian Ibarz}, \bibinfo{person}{Aaron Jaffey},
  \bibinfo{person}{Alek Jaworski}, \bibinfo{person}{Alexander Kaplan},
  \bibinfo{person}{Harshit Khaitan}, \bibinfo{person}{Andy Koch},
  \bibinfo{person}{Naveen Kumar}, \bibinfo{person}{Steve Lacy},
  \bibinfo{person}{James Laudon}, \bibinfo{person}{James Law},
  \bibinfo{person}{Diemthu Le}, \bibinfo{person}{Chris Leary},
  \bibinfo{person}{Zhuyuan Liu}, \bibinfo{person}{Kyle Lucke},
  \bibinfo{person}{Alan Lundin}, \bibinfo{person}{Gordon MacKean},
  \bibinfo{person}{Adriana Maggiore}, \bibinfo{person}{Maire Mahony},
  \bibinfo{person}{Kieran Miller}, \bibinfo{person}{Rahul Nagarajan},
  \bibinfo{person}{Ravi Narayanaswami}, \bibinfo{person}{Ray Ni},
  \bibinfo{person}{Kathy Nix}, \bibinfo{person}{Thomas Norrie},
  \bibinfo{person}{Mark Omernick}, \bibinfo{person}{Narayana Penukonda},
  \bibinfo{person}{Andy Phelps}, \bibinfo{person}{Jonathan Ross},
  \bibinfo{person}{Matt Ross}, \bibinfo{person}{Amir Salek},
  \bibinfo{person}{Emad Samadiani}, \bibinfo{person}{Chris Severn},
  \bibinfo{person}{Gregory Sizikov}, \bibinfo{person}{Matthew Snelham},
  \bibinfo{person}{Jed Souter}, \bibinfo{person}{Dan Steinberg},
  \bibinfo{person}{Andy Swing}, \bibinfo{person}{Mercedes Tan},
  \bibinfo{person}{Gregory Thorson}, \bibinfo{person}{Bo Tian},
  \bibinfo{person}{Horia Toma}, \bibinfo{person}{Erick Tuttle},
  \bibinfo{person}{Vijay Vasudevan}, \bibinfo{person}{Richard Walter},
  \bibinfo{person}{Walter Wang}, \bibinfo{person}{Eric Wilcox}, {and}
  \bibinfo{person}{Doe~Hyun Yoon}.} \bibinfo{year}{2017}\natexlab{}.
\newblock \showarticletitle{{In-Datacenter Performance Analysis of a Tensor
  Processing Unit}}.
\newblock \bibinfo{journal}{{\em arXiv preprint arXiv:1704.04760\/}}
  (\bibinfo{year}{2017}).
\newblock


\bibitem[\protect\citeauthoryear{Kingma, Salimans, Jozefowicz, Chen, Sutskever,
  and Welling}{Kingma et~al\mbox{.}}{2016}]%
        {kingma2016improved}
\bibfield{author}{\bibinfo{person}{Diederik~P Kingma}, \bibinfo{person}{Tim
  Salimans}, \bibinfo{person}{Rafal Jozefowicz}, \bibinfo{person}{Xi Chen},
  \bibinfo{person}{Ilya Sutskever}, {and} \bibinfo{person}{Max Welling}.}
  \bibinfo{year}{2016}\natexlab{}.
\newblock \showarticletitle{{Improving Variational Inference with Inverse
  Autoregressive Flow}}. In \bibinfo{booktitle}{{\em Neural Information
  Processing Systems}}.
\newblock


\bibitem[\protect\citeauthoryear{Kingma and Welling}{Kingma and
  Welling}{2014}]%
        {kingma2014auto}
\bibfield{author}{\bibinfo{person}{Diederik~P Kingma} {and}
  \bibinfo{person}{Max Welling}.} \bibinfo{year}{2014}\natexlab{}.
\newblock \showarticletitle{Auto-Encoding Variational {B}ayes}. In
  \bibinfo{booktitle}{{\em International Conference on Learning
  Representations}}.
\newblock


\bibitem[\protect\citeauthoryear{Kucukelbir, Tran, Ranganath, Gelman, and
  Blei}{Kucukelbir et~al\mbox{.}}{2017}]%
        {kucukelbir2017automatic}
\bibfield{author}{\bibinfo{person}{Alp Kucukelbir}, \bibinfo{person}{Dustin
  Tran}, \bibinfo{person}{Rajesh Ranganath}, \bibinfo{person}{Andrew Gelman},
  {and} \bibinfo{person}{David~M Blei}.} \bibinfo{year}{2017}\natexlab{}.
\newblock \showarticletitle{Automatic Differentiation Variational Inference}.
\newblock \bibinfo{journal}{{\em Journal of Machine Learning Research\/}}
  \bibinfo{volume}{18}, \bibinfo{number}{14} (\bibinfo{year}{2017}),
  \bibinfo{pages}{1--45}.
\newblock


\bibitem[\protect\citeauthoryear{Maddison, Mnih, and Teh}{Maddison
  et~al\mbox{.}}{2017}]%
        {maddison2017concrete}
\bibfield{author}{\bibinfo{person}{Chris~J. Maddison}, \bibinfo{person}{Andriy
  Mnih}, {and} \bibinfo{person}{Yee~Whye Teh}.}
  \bibinfo{year}{2017}\natexlab{}.
\newblock \showarticletitle{{The Concrete Distribution: A Continuous Relaxation
  of Discrete Random Variables}}. In \bibinfo{booktitle}{{\em International
  Conference on Learning Representations}}.
\newblock


\bibitem[\protect\citeauthoryear{Marsaglia and Tsang}{Marsaglia and
  Tsang}{2000}]%
        {marsaglia2000simple}
\bibfield{author}{\bibinfo{person}{George Marsaglia} {and}
  \bibinfo{person}{Wai~Wan Tsang}.} \bibinfo{year}{2000}\natexlab{}.
\newblock \showarticletitle{A simple method for generating gamma variables}.
\newblock \bibinfo{journal}{{\em ACM Transactions on Mathematical Software
  (TOMS)\/}} \bibinfo{volume}{26}, \bibinfo{number}{3} (\bibinfo{year}{2000}),
  \bibinfo{pages}{363--372}.
\newblock


\bibitem[\protect\citeauthoryear{Morimoto}{Morimoto}{1963}]%
        {morimoto1963markov}
\bibfield{author}{\bibinfo{person}{Tetsuzo Morimoto}.}
  \bibinfo{year}{1963}\natexlab{}.
\newblock \showarticletitle{Markov processes and the H-theorem}.
\newblock \bibinfo{journal}{{\em Journal of the Physical Society of Japan\/}}
  \bibinfo{volume}{18}, \bibinfo{number}{3} (\bibinfo{year}{1963}),
  \bibinfo{pages}{328--331}.
\newblock


\bibitem[\protect\citeauthoryear{M{\"u}ller}{M{\"u}ller}{1997}]%
        {muller1997integral}
\bibfield{author}{\bibinfo{person}{Alfred M{\"u}ller}.}
  \bibinfo{year}{1997}\natexlab{}.
\newblock \showarticletitle{Integral probability metrics and their generating
  classes of functions}.
\newblock \bibinfo{journal}{{\em Advances in Applied Probability\/}}
  \bibinfo{volume}{29}, \bibinfo{number}{2} (\bibinfo{year}{1997}),
  \bibinfo{pages}{429--443}.
\newblock


\bibitem[\protect\citeauthoryear{Murphy}{Murphy}{2012}]%
        {murphy2012machine}
\bibfield{author}{\bibinfo{person}{Kevin~P Murphy}.}
  \bibinfo{year}{2012}\natexlab{}.
\newblock \bibinfo{booktitle}{{\em Machine Learning: a Probabilistic
  Perspective}}.
\newblock \bibinfo{publisher}{MIT press}.
\newblock


\bibitem[\protect\citeauthoryear{Naesseth, Ruiz, Linderman, and Blei}{Naesseth
  et~al\mbox{.}}{2017}]%
        {naesseth-reparameterizable-rejection}
\bibfield{author}{\bibinfo{person}{Christian Naesseth},
  \bibinfo{person}{Francisco Ruiz}, \bibinfo{person}{Scott Linderman}, {and}
  \bibinfo{person}{David~M Blei}.} \bibinfo{year}{2017}\natexlab{}.
\newblock \showarticletitle{{Reparameterization Gradients through
  Acceptance-Rejection Sampling Algorithms}}. In \bibinfo{booktitle}{{\em
  Artificial Intelligence and Statistics}}.
\newblock


\bibitem[\protect\citeauthoryear{Niederreiter}{Niederreiter}{1978}]%
        {niederreiter1978quasi}
\bibfield{author}{\bibinfo{person}{Harald Niederreiter}.}
  \bibinfo{year}{1978}\natexlab{}.
\newblock \showarticletitle{Quasi-Monte Carlo methods and pseudo-random
  numbers}.
\newblock \bibinfo{journal}{{\it Bull. Amer. Math. Soc.}} \bibinfo{volume}{84},
  \bibinfo{number}{6} (\bibinfo{year}{1978}), \bibinfo{pages}{957--1041}.
\newblock


\bibitem[\protect\citeauthoryear{Papamakarios, Pavlakou, and
  Murray}{Papamakarios et~al\mbox{.}}{2017}]%
        {papamakarios2017masked}
\bibfield{author}{\bibinfo{person}{George Papamakarios}, \bibinfo{person}{Theo
  Pavlakou}, {and} \bibinfo{person}{Iain Murray}.}
  \bibinfo{year}{2017}\natexlab{}.
\newblock \showarticletitle{Masked Autoregressive Flow for Density Estimation}.
  In \bibinfo{booktitle}{{\em Neural Information Processing Systems}}.
\newblock


\bibitem[\protect\citeauthoryear{Peterson}{Peterson}{[n. d.]}]%
        {petersonsix}
\bibfield{author}{\bibinfo{person}{Benjamin Peterson}.} \bibinfo{year}{[n.
  d.]}\natexlab{}.
\newblock \bibinfo{title}{six: Python 2 and 3 compatibility utilities}.
\newblock \bibinfo{howpublished}{\url{https://github.com/benjaminp/six}}.
  (\bibinfo{year}{[n. d.]}).
\newblock


\bibitem[\protect\citeauthoryear{{Pyro Developers}}{{Pyro Developers}}{2017}]%
        {pyro2017pyro}
\bibfield{author}{\bibinfo{person}{{Pyro Developers}}.}
  \bibinfo{year}{2017}\natexlab{}.
\newblock \bibinfo{title}{Pyro}.
\newblock \bibinfo{howpublished}{\url{https://github.com/pyro/pyro}}.
  (\bibinfo{year}{2017}).
\newblock


\bibitem[\protect\citeauthoryear{{Pytorch Developers}}{{Pytorch
  Developers}}{2017}]%
        {pytorch2017pytorch}
\bibfield{author}{\bibinfo{person}{{Pytorch Developers}}.}
  \bibinfo{year}{2017}\natexlab{}.
\newblock \bibinfo{title}{Pytorch}.
\newblock \bibinfo{howpublished}{\url{https://github.com/pytorch/pytorch}}.
  (\bibinfo{year}{2017}).
\newblock


\bibitem[\protect\citeauthoryear{Ranganath, Gerrish, and Blei}{Ranganath
  et~al\mbox{.}}{2014}]%
        {ranganath2014black}
\bibfield{author}{\bibinfo{person}{Rajesh Ranganath}, \bibinfo{person}{Sean
  Gerrish}, {and} \bibinfo{person}{David Blei}.}
  \bibinfo{year}{2014}\natexlab{}.
\newblock \showarticletitle{Black box variational inference}. In
  \bibinfo{booktitle}{{\em Artificial Intelligence and Statistics}}.
\newblock


\bibitem[\protect\citeauthoryear{Salimans, Karpathy, Chen, and Kingma}{Salimans
  et~al\mbox{.}}{2017}]%
        {salimans2017pixelcnn++}
\bibfield{author}{\bibinfo{person}{Tim Salimans}, \bibinfo{person}{Andrej
  Karpathy}, \bibinfo{person}{Xi Chen}, {and} \bibinfo{person}{Diederik~P
  Kingma}.} \bibinfo{year}{2017}\natexlab{}.
\newblock \showarticletitle{{PixelCNN++}: Improving the PixelCNN with
  Discretized Logistic Mixture Likelihood and Other Modifications}. In
  \bibinfo{booktitle}{{\em International Conference on Learning
  Representations}}.
\newblock


\bibitem[\protect\citeauthoryear{Schulman, Heess, Weber, and Abbeel}{Schulman
  et~al\mbox{.}}{2015}]%
        {schulman2015gradient}
\bibfield{author}{\bibinfo{person}{John Schulman}, \bibinfo{person}{Nicolas
  Heess}, \bibinfo{person}{Theophane Weber}, {and} \bibinfo{person}{Pieter
  Abbeel}.} \bibinfo{year}{2015}\natexlab{}.
\newblock \showarticletitle{Gradient Estimation Using Stochastic Computation
  Graphs}. In \bibinfo{booktitle}{{\em Neural Information Processing Systems}}.
\newblock


\bibitem[\protect\citeauthoryear{Shi, Chen, Zhu, Sun, Luo, Gu, and Zhou}{Shi
  et~al\mbox{.}}{2017}]%
        {shi2017zhusuan}
\bibfield{author}{\bibinfo{person}{Jiaxin Shi}, \bibinfo{person}{Jianfei Chen},
  \bibinfo{person}{Jun Zhu}, \bibinfo{person}{Shengyang Sun},
  \bibinfo{person}{Yucen Luo}, \bibinfo{person}{Yihong Gu}, {and}
  \bibinfo{person}{Yuhao Zhou}.} \bibinfo{year}{2017}\natexlab{}.
\newblock \showarticletitle{ZhuSuan: A Library for {B}ayesian Deep Learning}.
\newblock \bibinfo{journal}{{\em arXiv preprint arXiv:1709.05870\/}}
  (\bibinfo{year}{2017}).
\newblock


\bibitem[\protect\citeauthoryear{Spivak}{Spivak}{[n. d.]}]%
        {spivak2010comprehensive}
\bibfield{author}{\bibinfo{person}{Michael Spivak}.} \bibinfo{year}{[n.
  d.]}\natexlab{}.
\newblock \showarticletitle{A Comprehensive Introduction to Differential
  Geometry, Vol. III}.
\newblock \bibinfo{journal}{{\em AMC\/}}  \bibinfo{volume}{10}
  (\bibinfo{year}{[n. d.]}), \bibinfo{pages}{12}.
\newblock


\bibitem[\protect\citeauthoryear{Tran and Blei}{Tran and Blei}{2017}]%
        {tran2017implicit}
\bibfield{author}{\bibinfo{person}{Dustin Tran} {and} \bibinfo{person}{David
  Blei}.} \bibinfo{year}{2017}\natexlab{}.
\newblock \showarticletitle{{Implicit Causal Models for Genome-wide Association
  Studies}}.
\newblock \bibinfo{journal}{{\em arXiv preprint arXiv:1710.10742\/}}
  (\bibinfo{year}{2017}).
\newblock


\bibitem[\protect\citeauthoryear{Tran, Hoffman, Saurous, Brevdo, Murphy, and
  Blei}{Tran et~al\mbox{.}}{2017}]%
        {tran2017deep}
\bibfield{author}{\bibinfo{person}{Dustin Tran}, \bibinfo{person}{Matthew~D
  Hoffman}, \bibinfo{person}{Rif~A Saurous}, \bibinfo{person}{Eugene Brevdo},
  \bibinfo{person}{Kevin Murphy}, {and} \bibinfo{person}{David~M Blei}.}
  \bibinfo{year}{2017}\natexlab{}.
\newblock \showarticletitle{{Deep Probabilistic Programming}}. In
  \bibinfo{booktitle}{{\em International Conference on Learning
  Representations}}.
\newblock


\bibitem[\protect\citeauthoryear{Tran, Kucukelbir, Dieng, Rudolph, Liang, and
  Blei}{Tran et~al\mbox{.}}{2016}]%
        {tran2016edward}
\bibfield{author}{\bibinfo{person}{Dustin Tran}, \bibinfo{person}{Alp
  Kucukelbir}, \bibinfo{person}{Adji~B Dieng}, \bibinfo{person}{Maja Rudolph},
  \bibinfo{person}{Dawen Liang}, {and} \bibinfo{person}{David~M Blei}.}
  \bibinfo{year}{2016}\natexlab{}.
\newblock \showarticletitle{{Edward: A library for probabilistic modeling,
  inference, and criticism}}.
\newblock \bibinfo{journal}{{\em arXiv preprint arXiv:1610.09787\/}}
  (\bibinfo{year}{2016}).
\newblock


\bibitem[\protect\citeauthoryear{van~den Oord, Dieleman, Zen, Simonyan,
  Vinyals, Graves, Kalchbrenner, Senior, and Kavukcuoglu}{van~den Oord
  et~al\mbox{.}}{2016a}]%
        {vandenoord2016wavenet}
\bibfield{author}{\bibinfo{person}{A{\"a}ron van~den Oord},
  \bibinfo{person}{Sander Dieleman}, \bibinfo{person}{Heiga Zen},
  \bibinfo{person}{Karen Simonyan}, \bibinfo{person}{Oriol Vinyals},
  \bibinfo{person}{Alex Graves}, \bibinfo{person}{Nal Kalchbrenner},
  \bibinfo{person}{Andrew Senior}, {and} \bibinfo{person}{Koray Kavukcuoglu}.}
  \bibinfo{year}{2016}\natexlab{a}.
\newblock \showarticletitle{{WaveNet: A Generative Model for Raw Audio}}.
\newblock \bibinfo{journal}{{\em arXiv preprint arXiv:1609.03499\/}}
  (\bibinfo{year}{2016}).
\newblock


\bibitem[\protect\citeauthoryear{van~den Oord, Kalchbrenner, and
  Kavukcuoglu}{van~den Oord et~al\mbox{.}}{2016b}]%
        {vandenoord2016pixel}
\bibfield{author}{\bibinfo{person}{Aaron van~den Oord}, \bibinfo{person}{Nal
  Kalchbrenner}, {and} \bibinfo{person}{Koray Kavukcuoglu}.}
  \bibinfo{year}{2016}\natexlab{b}.
\newblock \showarticletitle{Pixel recurrent neural networks}. In
  \bibinfo{booktitle}{{\em International Conference on Machine Learning}}.
\newblock


\bibitem[\protect\citeauthoryear{Walt, Colbert, and Varoquaux}{Walt
  et~al\mbox{.}}{2011}]%
        {walt2011numpy}
\bibfield{author}{\bibinfo{person}{St{\'e}fan van~der Walt},
  \bibinfo{person}{S~Chris Colbert}, {and} \bibinfo{person}{Gael Varoquaux}.}
  \bibinfo{year}{2011}\natexlab{}.
\newblock \showarticletitle{The NumPy array: a structure for efficient
  numerical computation}.
\newblock \bibinfo{journal}{{\em Computing in Science \& Engineering\/}}
  \bibinfo{volume}{13}, \bibinfo{number}{2} (\bibinfo{year}{2011}),
  \bibinfo{pages}{22--30}.
\newblock


\bibitem[\protect\citeauthoryear{Wasserman}{Wasserman}{2013}]%
        {wasserman2013all}
\bibfield{author}{\bibinfo{person}{Larry Wasserman}.}
  \bibinfo{year}{2013}\natexlab{}.
\newblock \bibinfo{booktitle}{{\em All of Statistics: A concise Course in
  Statistical Inference}}.
\newblock \bibinfo{publisher}{Springer Science \& Business Media}.
\newblock


\end{thebibliography}


\end{document}